\pdfoutput=1

\documentclass[11pt]{article}

\usepackage{EMNLP2022}

\usepackage{times}
\usepackage{latexsym}

\usepackage[T1]{fontenc}

\usepackage[utf8]{inputenc}

\usepackage{microtype}

\usepackage{inconsolata}

\usepackage{todonotes}
\usepackage{defs}
\usepackage{algorithm}
\usepackage{algorithmic}
\usepackage{amsmath}
\usepackage{multirow}
\usepackage{booktabs}
\usepackage{hyperref}
\newtheorem{corollary}{Corollary}
\newtheorem{lemma}{Lemma}
\newcommand{\vocab}{\mathcal{M}}
\newcommand{\prnnt}{P_\text{rnnt}}

\newcommand{\dsys}{\textsc{PGM}}
\newcommand{\model}{\textsc{RNN-T}}
\newcommand{\random}{{\em Random-Subset}}
\newcommand{\largeSmall}{{\em LargeSmall}}
\newcommand{\largeOnly}{{\em LargeOnly}}
\def\vy{{\mathbf{y}}}

\makeatletter
\def\thanks#1{\protected@xdef\@thanks{\@thanks
        \protect\footnotetext{#1}}}
\makeatother

\title{Partitioned Gradient Matching based Data Subset Selection \\ for Compute-Efficient \& Robust ASR Training}

\author{Ashish Mittal$^{1,2,\dagger}$, Durga Sivasubramanian$^{2,\dagger}$ \thanks{$\dagger$Equal contribution. Correspondence to: arakeshk@in.ibm.com, durgas@cse.iitb.ac.in}, Rishabh Iyer$^3$,\\ {\bf Preethi Jyothi}$^2$, {\bf Ganesh Ramakrishnan}$^2$   \\
         $^1$ IBM Research, India \\$^2$ Indian Institute of Technology Bombay, Mumbai, India \\ $^3$ The University of Texas at Dallas, Dallas, USA}

\begin{document}
\maketitle
\begin{abstract}
Training state-of-the-art ASR systems such as \model\ often have a high associated financial and environmental cost. Training with a subset of training data could mitigate this problem if the subset selected could achieve performance on-par with training with the entire dataset. Although there are many data subset selection (DSS) algorithms, direct application to the \model\ is difficult, especially the DSS algorithms that are adaptive and use learning dynamics such as gradients, since \model\ tends to have gradients with a significantly larger memory footprint. In this paper we propose \textbf{P}artitioned \textbf{G}radient \textbf{M}atching (\dsys) a novel distributable DSS algorithm, suitable for massive datasets like those used to train \model. Through extensive experiments on Librispeech 100H and Librispeech 960H, we show that \dsys\ achieves between $3\times$ to $6\times$ speedup with only a very small accuracy degradation (under $1\%$ absolute WER difference). In addition, we demonstrate similar results for \dsys\ even in settings where the training data is corrupted with noise.



\end{abstract}

\section{Introduction}

Owing to their simplicity in directly mapping an acoustic input sequence to a output sequence of characters, or words, or even word-pieces, neural end-to-end methods~\cite{graves2006connectionist,graves2013speech,chan2016listen,vaswani2017attention,he2019streaming} have become ubiquitous. The most common end-to-end architectures include (i) Connectionist Temporal Classification (CTC) models~\cite{graves2006connectionist,gulati2020conformer}, (ii) Attention-based Encoder-Decoder models (AED)~\cite{chan2016listen, watanabe2017hybrid} and (iii) Sequence Transduction models \cite{graves2012sequence} such as RNN-Ts~\cite{graves2013speech}. Due to their streaming and low-latency properties, sequence transduction architectures such as RNN-T~\cite{graves2013speech, sainath2020streaming,saon2021advancing} are becoming state-of-the-art for modeling the ASR problem. 

These successes in the ASR have come at a cost, as most of the practical RNN-T models are trained on thousands of hours of labeled datasets~\cite{rao2017exploring, zhao2021addressing}. Model training on these massive datasets leads to significantly increased training time, energy requirements, and consequently the carbon footprint~\cite{sharir2020cost, strubell2019energy, schwartz2020green, parcollet2021energy}.  As per Parcollet {\em et al.}~\cite{parcollet2021energy}, training an RNN-T model on Librispeech 960H~\cite{panayotov2015librispeech} emits more than 10kg $CO_{2}$  if trained in France, which becomes much worse for developing countries. This is exacerbated due to the many more training runs required for hyper-parameter tuning. This warrants a need for greener training strategies that rely on significantly lower resources while still achieving state-of-the-art results.
\vspace{-2mm}

One way to make ASR training more efficient is to train on a subset of the  training data, which ensures minimum performance loss~\cite{pmlr-v139-killamsetty21a,wei2014fast, kaushal2019learning,coleman2020selection,har2004coresets,clarkson2010coresets,mirzasoleiman2020coresets,killamsetty2021glister,liu2017svitchboard}. Since training on a subset reduces end-to-end time, the hyperparameter tuning time is also reduced. While greedy subset selection algorithms employ various criteria to identify the appropriate subset of training points, the process of forming the subsets remains sequential. 
However, for a large scale speech corpus such as  Librispeech~\cite{panayotov2015librispeech} this requirement may be difficult to meet. In this work, we propose a Partitioned Gradient Matching (\dsys) approach, which scales well with huge datasets used in ASR and takes advantage of distributed setups. To the best of our knowledge, this is the first such study performed for ASR systems.


\subsection{Contributions of this work}
\textbf{The \dsys{} Algorithm:} We present \dsys{} a data subset selection algorithm 
which constructs partial subsets from data partitions of the original dataset. This circumvents the need to load the entire dataset at a time into the memory, which is otherwise prohibitively expensive for ASR systems such as \model (see Section~\ref{need}). 

\noindent \textbf{\dsys{} is a distributable Algorithm:} Training with a subset of the training data is beneficial only when the cost of selecting a subset is also less. Therefore, for subset selection algorithms to scale to larger datasets used in speech recognition, they must work across multiple GPUs, since training for ASR systems can then be distributed. In Section~\ref{dist}, we present \dsys{} which is more suitable for ASR systems, more specifically for RNN-T.

\noindent \textbf{Trade-off between efficiency and accuracy:}
A subset selection algorithm has to counter the contrasting goals of efficiency and accuracy. We perform extensive experiments to demonstrate the trade-off between efficiency and accuracy for \dsys{} and provide a general recipe for a user to control the trade-off.

\noindent \textbf{Effectiveness of \dsys{} in a Noisy ASR setting:}
A subset selection algorithm should work well when the training data is corrupted with noise. In this work, we show the efficacy of  \dsys{}, even when a fraction of the labeled dataset is augmented with noise across varying signal-to-noise ratios.

\section{Background: RNN Transducer}

The RNN-T model~\cite{graves2013speech,graves2012sequence} maps an input acoustic signal $(x_1, x_2, \dots, x_T)$ to an output sequence $(y_1, y_2, \dots, y_U)$, where each output symbol $y_i \in \vocab$, and $\vocab$ is the vocabulary.  An RNN-T model consists of three components - (i) Transcription Network - which maps an acoustic signal $(x_1, x_2, \dots, x_T)$ to an encoded representation $(h_1, h_2, \dots, h_T)$,  $T$ being the length of the acoustic signal and $x_i$ being a $W$ dimensional feature representation, (ii) Prediction Network - which is a language model that maps the previously emitted non-blank tokens $\vy_{<U} = y_1, y_2, \dots, y_{u-1}$ to an output space $g_U$ for the next output token. (iii) Joint Network - that combines the Transcription Network representation $h_t$ and Prediction Network representation $g_u$ to produce $z_{t,u}$ using a feed-forward network $J$ and $\oplus$ as a combination operator (typically a sum).

\vspace{-.35cm}
\begin{equation}
\label{eq:prnnt1}
\begin{aligned}
    h_t = TranscriptionNetwork(x, t) 
\end{aligned}
\end{equation}

\vspace{-.35cm}
\begin{equation}
\label{eq:prnnt2}
\begin{aligned}
    g_u = PredictionNetwork(y, u)
\end{aligned}
\end{equation}


 During the training, the output probability $\prnnt(y_{t,u})$ over the output sequence $\vy$ is marginalized over all possible alignments using an efficient forward-backward algorithm to compute the log-likelihood. The training objective is to minimize the Negative Log Likelihood of the target sequence.
 
\vspace{-.35cm}
\begin{equation}
\label{eq:prnnt3}
\begin{aligned}
    \prnnt(y_{t,u} | \vy_{<u}, x_t) &=& \text{softmax}(J(h_t \oplus g_u))
\end{aligned}
\end{equation}

\vspace{-.35cm}
 \begin{equation}
\label{eq:prnnt4}
\begin{aligned}
    \mathcal{L} =  - ln Pr(y|x)
\end{aligned}
\end{equation}

 For inference, the decoding algorithms~\cite{graves2012sequence, saon2020alignment} attempt to find the best $(t, u)$ \def\vy{{\mathbf{y}}} and their corresponding output sequence $\vy$ using a beam search. In this work, we use the gradients of the joint network layer (J) for \dsys, since the linear layer helps in fusing the audio($h_t$) and the text($g_u$) representations.

\section{Limitations of existing subset selection algorithms}\label{need}

An approach to the selection of a subset of points from the entire dataset is to rank points based on their suitability. This ranking can be done either via a some static metric such as diversity or representation among features~\cite{wei2014fast, kaushal2019learning} or via a dynamic metric using instance-wise  loss gradients\footnote{gradient associated with an instance $(x,y)$ as opposed to mean mini-batch loss gradient used in training the model} to construct the subset greedily~\cite{mirzasoleiman2020coresets, killamsetty2021glister, pmlr-v139-killamsetty21a}.  In the latter case, ranking and re-ranking happens using instance-wise loss gradients. Specifically, during the selection process, loss gradients of the entire set of instances have to be available in the memory in order to perform greedy selection, since otherwise, subset selection time would be prohibitively large owing to disk reads, {\em etc}.

\begin{figure*}
\begin{center}
\includegraphics[width = 0.95\textwidth]{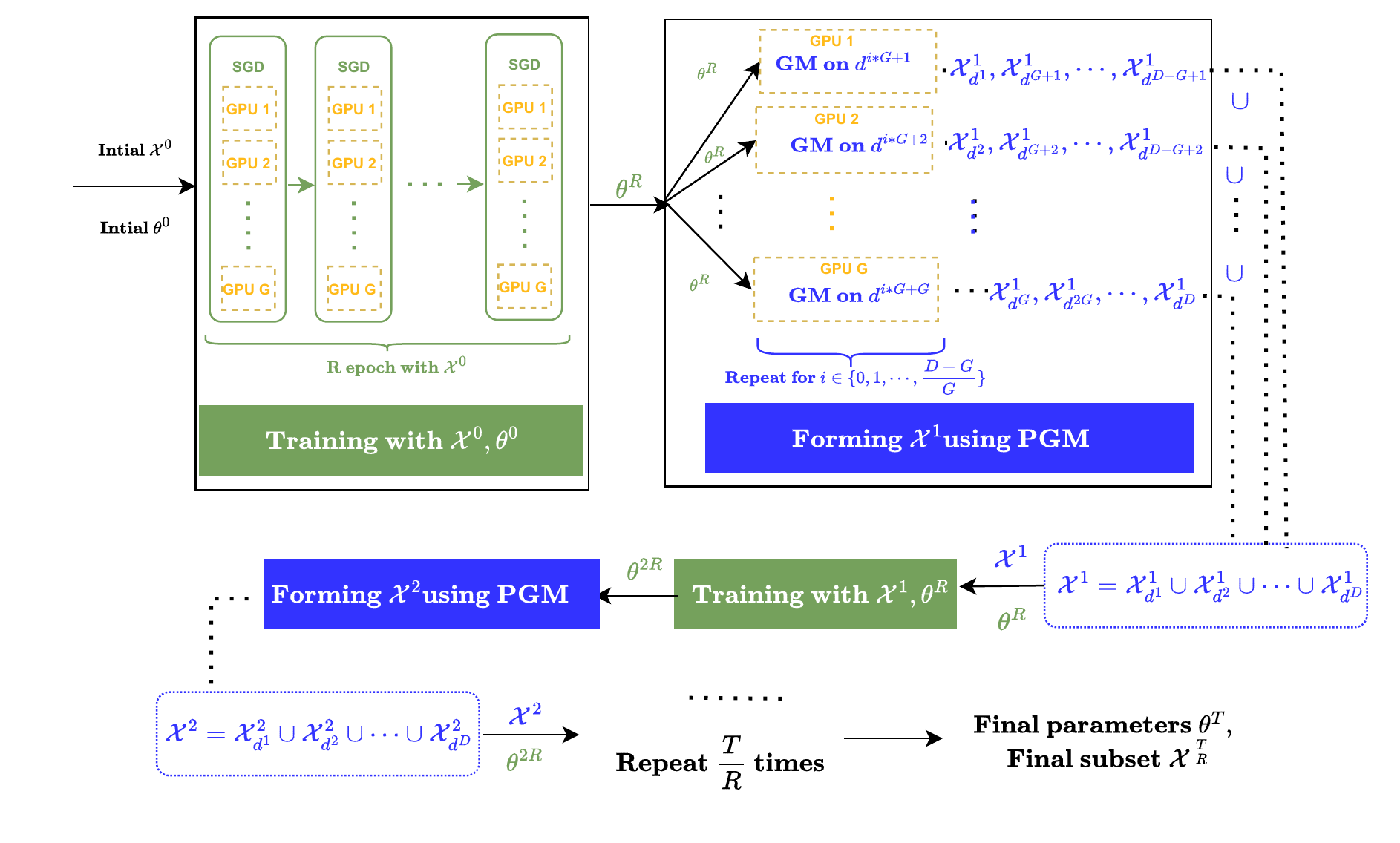}
\caption{As \dsys{} is a adaptive DSS algorithm, \dsys{} is invoked after for every $R$ epochs training \model\ using stochastic gradient descent. At every time step, using the latest set of parameters, \dsys{} forms partial subsets via Gradient Matching (GM) across GPUs. These partial subsets are combined and used for the next $R$ epochs of \model\ training. This is repeated  until the final set of parameters is obtained.}
\label{adaptivedss}
\end{center}
\end{figure*}

As keeping all the loss gradients in the memory would be resource intensive, we employ the following approximations, which have been also previously  employed by~\cite{mirzasoleiman2020coresets,killamsetty2021glister,pmlr-v139-killamsetty21a}, {\em viz.}, (i) only last layer gradients are used and (ii)  subsets are constructed for each class. The latter technique is not relevant in ASR systems since ASR requires sequential decoding into a large size vocabulary. Similar to the last layer approximation, for the \model{} model, we use the gradients of the joint network layer (J) which performs the important task of fusing speech ($h_t$) and text ($g_u$) features for sequence transduction. In Table~\ref{tab:mem_foot} we present the memory footprint of the last layer gradient obtained while training ResNet18~\cite{he2016deep} using CIFAR10~\cite{Krizhevsky09learningmultiple} and gradients of the joint network layer of \model\ using  Librispeech 100H. We compare against training ResNet18 using CIFAR10, since  most of these subset selection algorithms are applied to image classification settings. In the first column of the table \ref{tab:mem_foot}, we present the memory footprint of single instance's loss gradient. Clearly, the loss gradients used to train \model{} have a much higher footprint than the ones used in image classification setting. The CIFAR10 dataset has 50,000 instances and Librispeech 100H has 20539. In the second column, we present the total memory required to store all the instance-wise loss gradients.
The memory requirement for \model's loss gradients prohibitively huge.  
Thus, storing all the instance-wise loss gradients at once is not feasible for \model{} systems.

\begin{table}[t]
    \centering
    \begin{tabular}{c| c| c| c} \hline \hline
     \multirow{3}{2em}{Dataset }& {Single} & {Total} & {Per }   \\
     & {Gradient} & {size} & {Batch}    \\
  & {size (MB)} & (GB) &{size (GB)} \\ \hline 
 {\small{CIFAR10}}& 0.0215 & 1.049 & 0.0082 \\ [0.7ex]\hline 
 {\small{Librispeech 100H}}& 4.096 & 111 &  28 \\ [0.7ex]\hline 
 \hline 
    \end{tabular}
    \caption{Memory footprint of last layer gradient obtained while training ResNet18 using CIFAR10 and gradients of the joint network layer of \model\ using  Librispeech 100H. We use a batch size of 128 for CIFAR10 and 4 for Librispeech 100H.}
    \label{tab:mem_foot}
    
\end{table}

\citet{pmlr-v139-killamsetty21a} propose another technique, {\em viz.}, the {\em PerBatch} version, wherein one selects mini-batches (like used in SGD) instead of individual instances. Reduction in memory by using this technique is also not much for ASR systems such as \model, since batch size used here is often small. For example, the batch size employed for the CIFAR10 dataset is typically of 128, as  proposed by~\cite{he2016deep}  whereas the batch size is  4 for Librispeech 100H as used in the SpeechBrain~\cite{ravanelli2021speechbrain} Librispeech RNN-T recipe. We present the memory required to store all the batch-wise loss gradients in the third column of Table \ref{tab:mem_foot}. Although this requirement may seem satisfiable with some high end computing resource, however shown are the memory requirements to store the instance-wise loss gradients only. If we add other memory needs such as space to store \model{} model and space to process features and gradient computations, effectively one needs much larger GPU memory that the figures presented in Table \ref{tab:mem_foot}. These memory issues become even more pronounced while performing subset selection with Librispeech 960H.

Another problem with the existing subset selection algorithms is that they are sequential in nature. This doesn't allow the selection algorithm to enjoy the speedup achieved using state of the art techniques such as parallelizing across multiple GPUs etc. This may cause the subset selection algorithm to be a bottleneck while training \model\ with datasets of the scale of Librispeech. Therefore, there is a need to design an data subset algorithm that doesn't need all the loss gradient to form a subset and could be distributed across GPUs. 

\section{Partitioned Gradient Matching Algorithm}\label{dist} 

\noindent Let $\Ucal = \{(x_i, y_i)\}_{i=1}^{N}$ denote the set of training examples, and  $\Vcal = \{(x_j, y_j)\}_{j=1}^{M}$,  the validation set. Let $\theta$ denote the ASR system's parameters with $\theta^t$ as the ASR system's parameters at the $t^{th}$ epoch. The training loss associated with the $i^{th}$ instance is denoted by $L_T^i(\theta) = L_T(x_i, y_i, \theta) = - \ln Pr(y_i|x_i)$. We denote the validation loss by $L_V = -\sum_{i \in 
\Vcal} ln Pr(y_i|x_i)$. Let the training data be divided into $D$ partitions, {\em i.e.}, $\Ucal = d^1 \cup d^2 \cup \cdots \cup d^D$ where each partition comprises of $\frac{N}{D}$ instances. Let $B$ be the batch size, $b_n = N/B$ be the total number of mini-batches and $b_k = k/B$  the number of batches to be selected.

Let $L^{d^p}_T$ be the training loss associated with a data partition $d^p$ and  $\nabla_{\theta}L^{d^p}_T= \{\nabla_{\theta} L_T^{d^pB_1}(\theta^t), \cdots, \nabla_{\theta} L_T^{d^pB_{l}}(\theta^t)\}$ denote the set of mini-batch gradients associated with the data partition $d^p$, where $l = \frac{b_n}{D}$. Let $L^{b_n}_T$ denote the set of mini-batch gradients. For each data partition $d^p$, we wish to perform gradient matching (GM), by optimising the following problem, 
$$\underset{{\Xcal^t_{d^p} \subseteq d^p, |\Xcal^t_{d^p}| \leq \frac{b_k}{D}}}{\operatorname{argmin\hspace{0.7mm}}} 
\min_{\mathbf{w}^t_{d^p}} \mbox{E}_\lambda(\mathbf{w}^t_{d^p}, \Xcal^t_{d^p}, L^{d^p}_T,\nabla_{\theta} L^{d^p}_T, \theta^t)$$
where,
\vspace{-.25cm}
\begin{equation}
    \begin{aligned}
     \mbox{E}_\lambda(\mathbf{w}^t_{d^p}&, \Xcal^t_{d^p}, L^{d^p}_T,\nabla_{\theta} L^{d^p}_T, \theta^t) = \lambda \lVert \mathbf{w}^t_{d^p} \rVert^2 + \\& \lVert \sum_{i \in \Xcal^t_{d^p}} \mathbf{w}^t_{id^p} \nabla_{\theta} L^{d^pB_i}_T - \nabla_{\theta} L^{d^p}_T(\theta^t) \rVert
    \end{aligned}
    \label{main_train}
\end{equation}

This selects a subset of batches $\Xcal^t_{d^p}$ and associated weights $\mathbf{w}^t_{d^p}$, such that the weighted sum of loss gradients associated with each instance in the subset are the best approximation of the loss gradient of the entire data partition $d^p$ while honoring the budget constraints. We perform gradient matching on mini-batch wise loss gradients only as it helps in reducing the memory needs. Similarly, we can define gradient matching problem with loss associated with the validation set as,

\vspace{-.15cm}
$$\underset{{\Xcal^t_{d^p} \subseteq d^p, |\Xcal^t_{d^p}| \leq \frac{b_k}{D}}}{\operatorname{argmin\hspace{0.7mm}}} 
\min_{\mathbf{w}^t_{d^p}} \mbox{E}_\lambda(\mathbf{w}^t_{d^p}, \Xcal^t_{d^p}, L_V,\nabla_{\theta} L^{d^p}_T, \theta^t)$$
where,
\vspace{-.25cm}
\begin{equation}
    \begin{aligned}
     \mbox{E}_\lambda(\mathbf{w}^t_{d^p},& \Xcal^t_{d^p}, L_V,\nabla_{\theta} L^{d^p}_T, \theta^t) = \lambda \lVert \mathbf{w}^t_{d^p} \rVert^2 + \\& \lVert \sum_{i \in \Xcal^t_{d^p}} \mathbf{w}^t_{id^p} \nabla_{\theta} L^{d^pB_i}_T - \nabla_{\theta} L_V(\theta^t) \rVert
    \end{aligned}
    \label{main_val}
\end{equation}

The optimization problem given in Eq.\eqref{main_train} is weakly submodular \cite{pmlr-v139-killamsetty21a,natarajan1995sparse}. Hence, we can effectively solve it using a greedy algorithm with approximation guarantees – we use orthogonal matching pursuit (OMP) algorithm~\cite{elenberg2018restricted} to find the subset and their associated weights. We also add to Eq.\eqref{main_train} an $l_2$ regularization component to discourage large weight assignments to any of the instances selected in the subset, thereby preventing the model from overfitting
on some samples.

\begin{algorithm}[!t]
\caption{\dsys: \textbf{P}artitioned \textbf{G}radient \textbf{M}atching }
\begin{algorithmic}
\REQUIRE Train set: $\Ucal = d^1 \cup d^2 \cup \cdots \cup d^D$ consisting of $D$ partitions; validation set: ${\mathcal V}$; initial subset: $\Xcal^{0}$; subset size: $b_k$; TOL: $\epsilon$; initial params: $\theta^{0}$; learning rate: $\alpha$; total epochs: $T$, selection interval: $R$, Validation Flag: \mbox{Val}, Batchsize: $B$
\FOR {epochs $t$ in $1, \cdots, T$}
    \IF{$(t \mbox{ mod } R == 0)$}
        \STATE $\Xcal^{t} = \phi, \mathbf{w}^{t}  = []$
        \FOR {data partition $p$ in $d^1, \cdots, d^D$}
            \IF{$\mbox{Val}$}
                \STATE $\Xcal_d^{t}, \mathbf{w}_d^t = \operatorname{GM}(L_V, \nabla_{\theta} L^{d^p}_T,\theta^{t}, \frac{b_k}{D}, \epsilon)$
            \ELSE
                \STATE $\Xcal_d^{t}, \mathbf{w}_d^t = \operatorname{GM}(L^{d^p}_T, \nabla_{\theta} L^{d^p}_T, \theta^{t}, \frac{b_k}{D}, \epsilon)$ 
            \ENDIF
        \STATE $\Xcal^{t} = \Xcal^{t} \cup \Xcal_d^{t}$ 
        \STATE Extend $\mathbf{w}^{t}$ with $\mathbf{w}_d^t$
        \ENDFOR
    \ELSE
        \STATE $\Xcal^{t} = \Xcal^{t-1}$
    \ENDIF
    \STATE $\theta_{t+1} = \mbox{BatchSGD}(\Xcal^{t}, \mathbf{w}^t, \alpha, B)$
\ENDFOR
\STATE Output final model parameters $\theta^{T}$
\end{algorithmic}
\label{alg:dgrad-match}
\end{algorithm}

\begin{algorithm}[t]
\caption{Gradient Matching (GM) }
\label{alg:algorithm1_sub}
\begin{algorithmic}
\REQUIRE Loss of the entire dataset(train or validation) : $L$, set
of mini-batch gradients $\nabla_{\theta} L^{B}_T$, current parameters $\theta^t$, budget $k$, TOL: $\epsilon$; 

\STATE $\Xcal = \phi, \Xcal_f =\phi, r = \nabla_{\theta} L$

\FOR {$|\Xcal| \leq k$ or $ \mbox{E}_\lambda(w, \Xcal, L,\nabla_{\theta} L^{B}_T, \theta^t) > \epsilon$}
\STATE Pick a element $j$ in $\nabla_{\theta} L^{B}_T$ which a maximum alignment with $r$
\STATE $\Xcal = \Xcal \cup j$
\STATE $\Xcal_f = \Xcal \cup $ \{set of instances in the batch $j$\}
\STATE Update $w = \min_{w} \mbox{E}_\lambda(w, \Xcal, L,\nabla_{\theta} L^{B}_T, \theta^t)$

\STATE Update $r =  r - \mbox{E}_\lambda(w, \Xcal, L,\nabla_{\theta} L^{B}_T, \theta^t)$
    
\ENDFOR
\STATE Return $\Xcal_f, w$ 
\end{algorithmic}
\end{algorithm}

The complete algorithm is presented in Algorithm~\ref{alg:dgrad-match}. In the algorithm, `Val' is a boolean flag that indicates whether to match the subset loss gradient with validation set loss gradient like in noisy settings (`Val=True') or with training set loss gradient (`Val=False'). Depending on the choice of the loss gradient, we perform gradient matching  with $L^{d^p}_T$, current model parameters $\theta_{t}$, budget $\frac{b_k}{D}$, and a stopping criterion $\epsilon$. We describe gradient matching in details in Algorithm \ref{alg:algorithm1_sub}. Once the appropriate batch for selection is determined, we form $\Xcal_f$ adding all the samples constituting the selected mini-batch. The model is then trained using the mini-batch SGD. We randomly shuffle elements in the subset $\Xcal^t$, divide them up into mini-batches of size $B$, and run mini-batch SGD with instance weights.

The complete block diagram of \dsys{} is presented in Figure~\ref{adaptivedss}. As the subset selection process is dependant on the model parameters, we repeat the subset selection every $R$ epochs. For each data partition $d^p$, we perform gradient matching (GM) individually and obtain partial subsets $\Xcal^t_{d^p}$, sequentially, one after another. However in the presence of multi-GPU settings, since the gradient matching within a a data partition can be performed independently from gradient matching in other data partitions, the gradient matchings could be executed in parallel. This allows one to take advantage of multi-GPU settings which is critical to efficiently process large datasets typically used to train \model. In Figure \ref{adaptivedss}, we illustrate parallelization of \dsys{} on the system with $G$ GPUs. Here, every $G$ partial subsets are obtained in parallel and this process is repeated $\frac{D}{G}$ times.
 
\subsection{Connection to existing work} \label{conn_ew}


In this section we discuss the connection of \dsys{} with \textsc{Grad-MatchPB} \cite{pmlr-v139-killamsetty21a} where subset is selected via solving the following problem, 

\vspace{-.15cm}
$$\underset{{\Xcal^t \subseteq \Ucal, |\Xcal^t| \leq b_k}}{\operatorname{argmin\hspace{0.7mm}}} 
\min_{\mathbf{w}^t} \mbox{E}_\lambda(\mathbf{w}^t, \Xcal^t, L, L^{b_n}_T, \theta^t)$$
where 
\vspace{-.25cm}
\begin{equation*}
    \begin{aligned}
     \mbox{E}_\lambda(\mathbf{w}^t, \Xcal^t,& L, L^{b_n}_T, \theta^t) = \lambda \lVert \mathbf{w}^t \rVert^2 + \\& \lVert \sum_{i \in \Xcal^t}\mathbf{w}^t_{i} \nabla_{\theta} L_T^{B_i}(\theta^t) - \nabla_{\theta} L(\theta^t) \rVert
    \end{aligned}
\end{equation*}

\begin{figure*}[!tb]
\centering
\begin{minipage}{\columnwidth}
  \centering
  \includegraphics[width = \columnwidth,height=5cm]{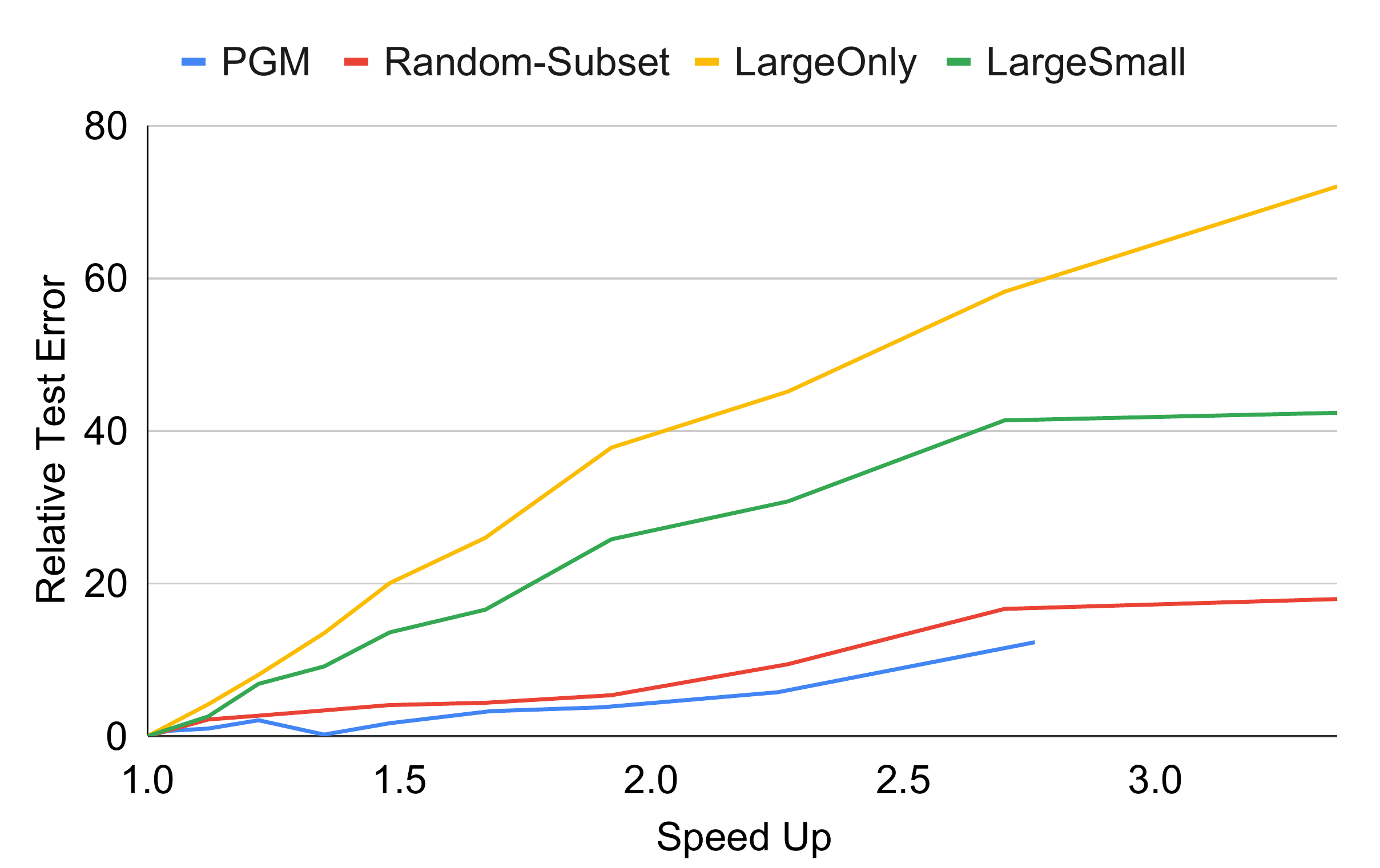}
  \captionof{figure}{\centering Relative Test Error($\downarrow$) {\em vs.} Speed Up($\uparrow$) for subset selection methods on Librispeech 100H \textsc{test-clean} test set. }
  \label{fig_1:wer}
  \vspace{-.3cm}
\end{minipage}%
\hfill
\begin{minipage}{\columnwidth}
  \centering
  \includegraphics[width = \columnwidth,height=5.5cm]{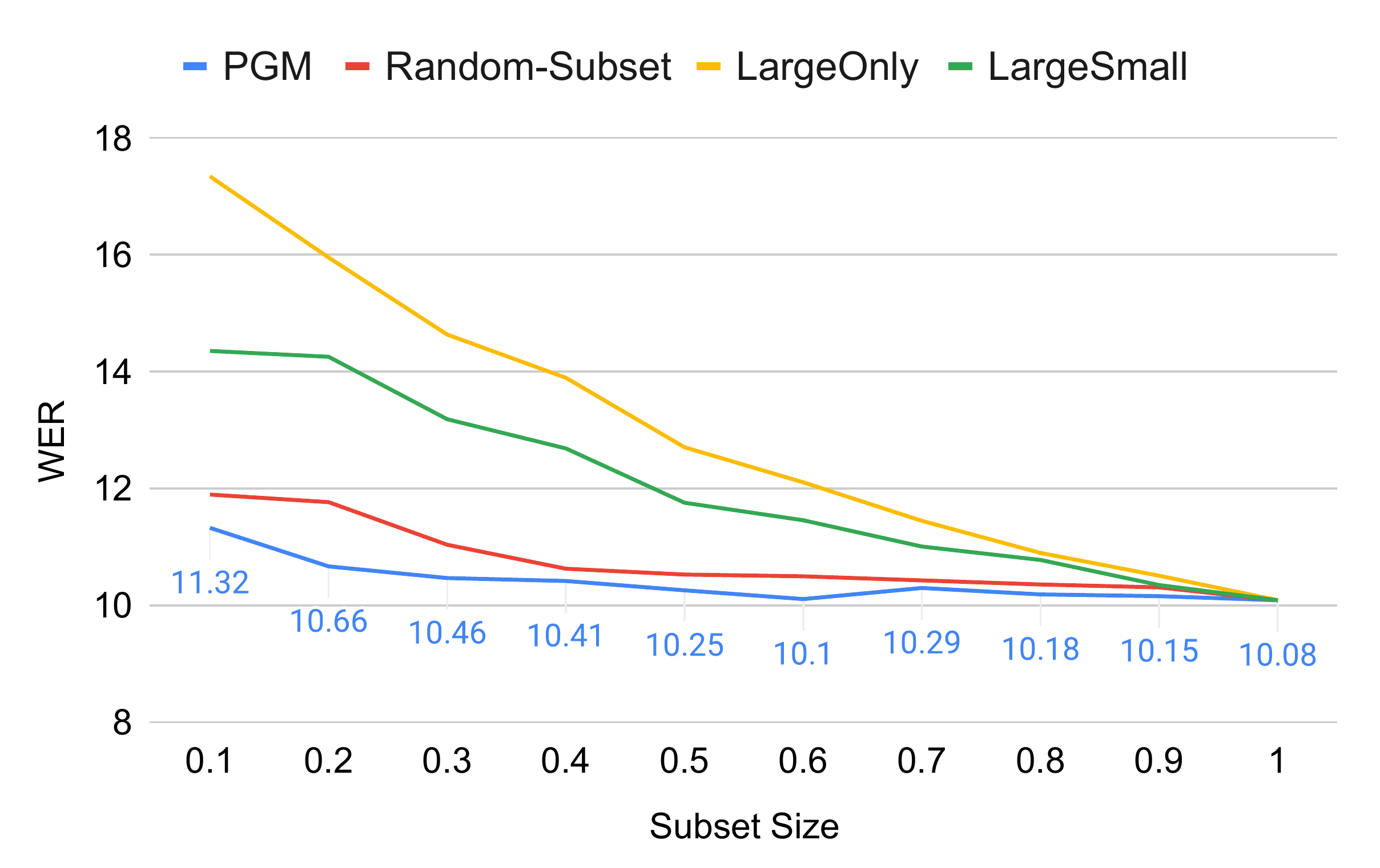}
  \captionof{figure}{Word Error Rate (WER) on the \textsc{test-clean} test set of Librispeech 100H for all the methods.}
  \label{fig_2}
  \vspace{-.3cm}
\end{minipage}
\end{figure*}

$L^{b_n}_T $ denotes the set of all mini-batch gradients, defined as $L^{b_n}_T = \nabla_{\theta} L^{d^1}_T \cup \nabla_{\theta} L^{d^2}_T\cup \cdots \cup \nabla_{\theta} L^{d^D}_T$ and $L$ is either the training loss of the entire dataset $L_T$ defined as $L_T = \mathbb{E}(L^{d^p}_T)$ or $L_V$ depending on what sort matching we seek for. The problem tries to find subset and it associated weights so that the gradients of the mini-batches best approximate the either gradient associated with the full dataset or the validation set. We show that \textsc{Grad-MatchPB} is lower bound to \dsys, that is 

\begin{equation*}
    \begin{aligned}
    \mathbb{E}(\mbox{E}_\lambda(\mathbf{w}^t_{d^p}, \Xcal^t_{d^p}, L^{d^p}_T,& \nabla_{\theta} L^{d^p}_T, \theta^t)) \\& \geq \mbox{E}_\lambda(\mathbf{w}^t, \Xcal^t, L_T, L^{b_n}_T, \theta^t))
    \end{aligned}
\end{equation*}
\vspace{-2mm}
and 
\vspace{-2mm}
\begin{equation*}
    \begin{aligned}
    \mathbb{E}(\mbox{E}_\lambda(\mathbf{w}^t_{d^p}, \Xcal^t_{d^p}, L_V,& \nabla_{\theta} L^{d^p}_T, \theta^t)) \\& \geq \mbox{E}_\lambda(\mathbf{w}^t, \Xcal^t, L_V, L^{b_n}_T, \theta^t))
    \end{aligned}
\end{equation*}

\noindent For the proof, we refer the reader to Appendix \ref{sec:proof}.

\section{Experiments} 

 \textbf{Datasets} We perform all our experiments on the Librispeech dataset~\cite{panayotov2015librispeech}. We present results on the medium-scale Librispeech 100H as well as on the large-scale Librispeech 960H datasets.

 Along with the standard Librispeech benchmark, we also perform experiments on noisy Librispeech, where the speech is augmented with noise across varying signal-to-noise ratios (up to 15db) on a fraction of the training data. We refer to this dataset as \textbf{Librispeech-noise}, where up to 30\% examples in the original dataset are augmented with noise across varying signal-to-noise ratios.
 
 \textbf{Architecture.} We perform all our experiments on the Speechbrain's\cite{ravanelli2021speechbrain} Librispeech transducer recipe. The transcription network of the RNN-T consists of a CRDNN encoder which has 2 CNN blocks followed by 4 layers of bi-LSTMs and subsequently followed by 2 DNN layers. The prediction network consists of an embedding layer followed by a single layer GRU unit. A joint network is a single linear layer that projects 1024 dimensional representations to output a vocabulary of 1000 BPE. The decoding is done through a time-synchronous decoding algorithm~\cite{graves2012sequence, hannun2019sequence} with a beam size of 4. The decoding involves an external transformer language model trained on the Librispeech corpus~\cite{kannan2018analysis,hrinchuk2020correction, wolf2019huggingface}.



\begin{figure*}[!tb]
\centering
\begin{minipage}{.8\textwidth}
\resizebox{\columnwidth}{!}{
\begin{tabular}{cl|ll|c}
\hline
\multicolumn{1}{l}{Subset} & Method                & \multicolumn{2}{c}{WER (Rel. Test Error) ($\downarrow$)} & Speed Up ($\uparrow$)             \\ \hline
\multicolumn{1}{l}{}            &                       & \textsc{test-clean}          & \textsc{test-other}          &                       \\ \hline \hline
\multicolumn{1}{l}{100\%}   & \multicolumn{1}{c|}{-} & 4.21 (0.0)          & 11.59 (0.0)         & \multicolumn{1}{c}{-} \\ \hline 
\multirow{2}{*}{10\%}           & \random          & 5.87 (39.43\%)      & 15.39 (32.79\%)     & 6.25                  \\
                                & \dsys            & 5.71 (35.63\%)      & 14.66 (26.49\%)     & 4.43                  \\ \hline \hline
\multirow{2}{*}{20\%}           & \random          & 5.08 (20.67\%)      & 13.89 (19.84\%)     & 3.95                  \\
                                & \dsys            & 5.01 (19\%)         & 12.78 (10.27\%)     & 3.30                  \\ \hline \hline
\multirow{2}{*}{30\%}           & \random         & 4.62 (12.65\%)      & 12.65 (9.74\%)      & 2.89                  \\
                                & \dsys            & 4.58 (8.79\%)       & 12.45 (7.42)        & 2.64                  \\ \hline
\end{tabular}}
\captionof{table}{\label{tab:wer_960}Results showing WER (Relative Test Error) and Speed Up on \textsc{test-clean} and \textsc{test-other} test splits of Librispeech 960H.}
\end{minipage}%
\hfill
\begin{minipage}{.9\textwidth}
  \centering
  \resizebox{\columnwidth}{!}{
\begin{tabular}{ll||cc||cc||cc}
\hline
                      & Subset & \multicolumn{2}{c}{Noise = 10\%} & \multicolumn{2}{c}{Noise = 20\%} & \multicolumn{2}{c}{Noise = 30\%} \\ \hline
                      &        & \random    & \dsys    & \random     & \dsys    & \random     & \dsys    \\ \hline \hline
\multirow{4}{*}{100H} & 100\%  & \multicolumn{2}{c||}{10.59}        & \multicolumn{2}{c||}{11.16}        & \multicolumn{2}{c}{11.39}        \\ 
                      & 10\%   & 11.79            & 11.86         & 11.64            & 11.67         & 11.96            & 11.82         \\
                      & 20\%   & 11.53            & 10.8          & 11.27            & \textbf{11.12}         & 11.39            & 11.25         \\
                      & 30\%   & 11.33            & \textbf{10.7}          & 11.74            & 11.42         & 12.05            & \textbf{11.17}         \\ \hline
\multirow{4}{*}{960H} & 100\%  & \multicolumn{2}{c||}{4.52}         & \multicolumn{2}{c||}{4.65}         & \multicolumn{2}{c}{4.68}             \\ 
                      & 10\%   & 6.5              & 6.28          & 6.44             & 6.54          & 6.58             & 6.43          \\
                      & 20\%   & 5.61             & 5.58          & 5.44             & 5.65          & 5.84             & 5.5           \\
                      & 30\%   & 4.99             & \textbf{4.97}          & 5.16             & \textbf{5.02}          & 5.62             &  \textbf{5.17} \\ \hline   
\end{tabular}}
\captionof{table}{\label{tab:wer_noise}Results showing WER on \textsc{test-clean} test set of  Librispeech 100H trained using noisy Librispeech dataset using \dsys{} and \random.}
\end{minipage}
\end{figure*}

 \textbf{Training Details.} For the training, we employ a learning rate of 2.0 with an annealing factor of 0.8 for the relative improvement of 0.0025 on validation loss (sometimes referred to as newbob scheduler). The training on Librispeech 100H is performed on two A100 40GB GPUs with the effective batch size of 8, whereas for Librispeech 960H, we employ two A100 80GB GPUs with an effective batch size of 24. All the training is done for 30 epochs. In all our experiments, the \dsys{} algorithm is invoked after every $5^{th}$ epoch ($R = 5$) after performing warm-start (training on full data) for 7 and 2 epochs on Librispeech 100H and Librispeech 960H datasets respectively. The results for each setting are averaged over 3 runs with different random seeds.

\textbf{\dsys{} Details.} For doing the subset selection with \dsys{}, we use the gradients of the Joint Network parameters, which we believe would have the maximum information concentrated for the sequence. We freeze the rest of the network while we compute the gradient of the Joint Network of the RNN-T. We use $D=7$ and $D=50$ (data partitions) to obtain subsets using the \dsys{} algorithm over gradients of training data for Librispeech 100H and 960H datasets respectively. Subset selection is performed using \textbf{{\em training set loss gradients}} in experiments performed using Librispeech 100H (Figures~\ref{fig_1:wer},\ref{fig_2}) and Librispeech 960H (table \ref{tab:wer_960}). For experiments with \textbf{Librispeech-noise} (Table~\ref{tab:wer_noise}) we employ the \textbf{{\em validation gradients}} for performing the subset selection, {\em since we are also concerned with robustness in the presence of noise}. 

\textbf{Baselines.} We compare the results obtained using the \dsys{} method against three intuitive baselines - (i) \textbf{\random{}} baseline, in which the subset of the dataset is obtained by choosing points with uniform probability. (ii) \textbf{\largeOnly{}} - For each subset, we employ only the largest utterances based on duration. (iii) \textbf{\largeSmall{}} - For each subset size, half of the subset is filled with smallest utterances and the other half with the largest utterances based on duration, to remove the length bias of the \textbf{\largeOnly{}} baseline. 

\subsection{Results}

To compare the efficacy of the \dsys, we compare the word error rate (WER), relative test error, and speed-up compared to  training with the entire dataset. We compute these metrics for both the Librispeech 100H and Librispeech 960H benchmarks.
Additionally, we also, present energy ratios {\em vs.} relative test error rate tradeoff on Librispeech 100H.

In Figure \ref{fig_1:wer}, we present the comparison of WER for \dsys{} against various baselines for various subset sizes of the full dataset. With just 20\% of the subset size, the \dsys{} method yields a WER of 10.66 as opposed to 10.08 obtained by training on the full dataset. For Librispeech 100H, \dsys{} consistently outperforms all the baseline, thus illustrating the effect 
of selecting subsets using the gradient matching algorithm. Also note, \random{} baseline is consistently better than other heuristic based baselines such LargeOnly and LargeSmall. In Figure~\ref{fig_2}, we plot the speed up against the Relative Test Error for Librispeech 100H. While \random{}  baseline is observed to attain higher speed up in comparison to the \dsys{} because of the simple selection strategy, \random{} baseline also incurs higher relative test error in comparison to the \dsys. 

\begin{figure}[!t]
\includegraphics[width = \columnwidth,height=5.5cm]{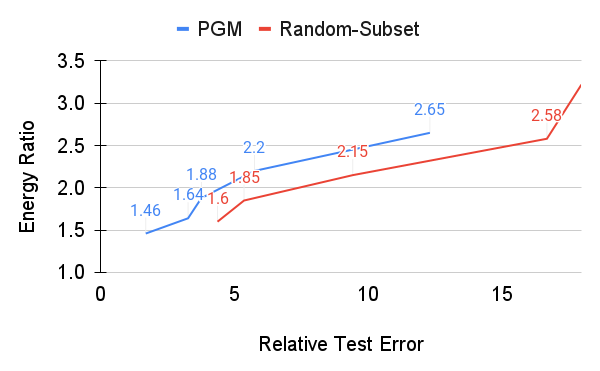}
\caption{Energy Ratio($\uparrow$) {\em vs.} Relative Test Error($\downarrow$) for \dsys{} and \random{} on Librispeech 100H.}
\label{fig4:energy}
\end{figure}

In Figure \ref{fig4:energy}, we present the plot of relative test error {\em w.r.t} energy efficiency for the full training setting. We use pyJoules\footnote{https://pypi.org/project/pyJoules/} for measuring the energy consumed by GPU cores.  We show that with \dsys, the training time is halved and energy efficiency is doubled while incurring the relative test error of less than 5\% as compared to the training on the entire dataset. For higher speedups, where there is a degradation in the WER, the loss is relatively better for \dsys{} as compared to the baseline. We do not show the energy efficiency for LargeOnly and LargeSmall baselines as their relative test error is consistently poor as compared to the Random Subset baseline as shown in Figure~\ref{fig_2}.

\begin{figure*}[!tb]
\centering
\begin{minipage}{\columnwidth}
\begin{tabular}{l|c|c}
\hline
                    & \random  & \multicolumn{1}{l}{\dsys} \\ 
                    \hline \hline
                
Overlap Index       & 20.2\%       & 6.37\%                         \\
Noise Overlap Index & 0.82\%        & 0.83\%               \\ \hline    \hline     
\end{tabular}
\captionof{table}{\label{tab:ablation-1} Overlap Indices - measures the overlap between consecutive subsets for \dsys{} and \random{} methods.}
\end{minipage}%
\hfill
\begin{minipage}{\columnwidth}
  \centering
  \begin{tabular}{c|c|c}
\hline 
Subset Size & WS = 2 epochs & WS= 3 epochs \\ \hline \hline
10\%        & 5.71                  & 5.3                   \\
20\%        & 5.01                  & 4.82                  \\
30\%        & 4.58                  & 4.54                 \\ \hline \hline
\end{tabular}
\captionof{table}{\label{tab:ablation-2} Effect of warm-start (WS) on WER for PGM on \textsc{test-clean} test set for Librispeech 960H}
\end{minipage}
\end{figure*}

\begin{table}[!t]
\begin{tabular}{l|c|c|c}
\hline
Subset Size       & nGPU = 1 & nGPU=2 & nGPU=2 \\ 
 & LR = 1.0 & LR = 1.0 & LR = 2.0 \\
                    \hline \hline
                
0.1  & 11.26      & 13.99     & 11.32           \\
0.2 & 10.6        & 12.58    &  10.66      \\ 
0.3 & 10.4 & 11.58 & 10.46 \\\hline \hline     
\end{tabular}
\captionof{table}{\label{tab:ablation-3} Effect of Learning Rate on WER for \dsys{} on \textsc{test-clean} test set of Librispeech 100H.}
\end{table}

For the ASR task we recommend using at least 30\% of the dataset for training the model or using more warm-start epochs as described in Section~\ref{subsec:ablation}.
In Table \ref{tab:wer_960}, we present comparison of the \dsys{} method with the baseline for the Librispeech 960H dataset on both the \textsc{test-clean} and \textsc{test-other} test sets. As shown in the Table, with just 30\% of the training data, \dsys{} is within 10\% of the relative test error (1\% of absolute error difference) when compared against training on the full data, thus yielding a speedup of 2.64. Similar results hold on the challenging \textsc{test-other} test set of the Librispeech which shows the better generalization of \dsys{} in comparison to the \random{} baseline. 

 \textbf{Results on Librispeech-noise}: We augment randomly selected signals from the dataset with noise across varying signal-to-noise ratios to mimic a more practical setting where subset selection algorithms need to address the noise while selecting useful subsets. We show the results on the Librispeech-noise 100H and 960H datasets for different subsets in Table \ref{tab:wer_noise}. \dsys{} consistently outperforms the \random{} baseline
for different subsets with lower relative test error when compared against the full training and still yields significant speed up to reduce training time and maintain robustness.

\subsection{Ablation Study}
\label{subsec:ablation}

Next, we do an ablation study to understand the effect of learning rate on \dsys{} for Librispeech 100H dataset. Since, the goal of subset selection algorithms is to reduce the training data for training, the older recipes (especially learning rate) on full training data do not work  as-is for the \dsys{} because of the distributed nature of the training. 

In Table \ref{tab:ablation-3}, we show the effect of learning rate on multi-gpu training of the \dsys{} method. The recipe for single GPU borrowed as-is for the multi-gpu training setting, performed poorly because the number of gradient updates in the distributed setting halved. To overcome this barrier, we doubled the learning rate to take larger steps and reach convergence within the same number of epochs.

We perform some ablation studies to understand why subsets selected by \dsys{} tend to outperform a relatively simple \random{} baseline. We compute the following two metrics:

 \textbf{Overlap Index (OI)}: This is the fraction of common points selected in the last two subset selection rounds with the subset size. This metric computes the diversity of the points being selected by the methods in the subsequent subset selection rounds. 

 \textbf{Noise Overlap Index (NOI)}: This is the fraction of noise points selected by the subset selection methods divided by the total number of noisy points. Both the metrics are computed by averaging the index for all the runs with the same subset selection method.

                

 As shown in Table \ref{tab:ablation-1}, \dsys{} selects more diverse points across different subset selection rounds which explains the better generalization of the \textsc{test-other} test set. At the same time, both the methods select a similar amount of noisy points during the subset selection indicating that \dsys{} selects more diverse points from the non-noisy points.

 Finally, we study the effect of warm-start on the performance of the \dsys{} algorithm. Since it is an adaptive data selection algorithm, \dsys{}  needs a good starting point for computing reasonable estimates of the gradients for subset selection. Table \ref{tab:ablation-2} shows the effect of warm-start epoch ablation on the \textsc{test-clean} test set for Librispeech 960H. As we increase the warm start, the performance of the \dsys{} algorithm improves at the cost of  speed up. 


\subsection{Comparing \dsys{} and \textsc{Grad-MatchPB}}

\begin{table*}[t]
    \centering
    \begin{tabular}{c| c| c| c| c| c} \hline 
     Subset-Size  & \random{} 	   & \largeSmall{} & \largeOnly & \textsc{Grad-MatchPB} & \dsys{}\\ \hline \hline
     
    0.1      & 16.64         & 17.98      &  17.27  & 16.14        & 16.23   \\
    0.2  & 16.43 & 17.23 &  16.35 & 15.89   & 16.03   \\
    0.3  & 16.28  & 16.35     &  16.22  & 15.79  & 15.95\\ \hline \hline
    \end{tabular}
    \caption{Comparison of WER obtained with  \random{}, \largeSmall{}, \largeOnly{},\textsc{Grad-MatchPB} and \dsys{} on TIMIT Phone recognition dataset.}
    \label{tab:pb_vs_pgm}
    
\end{table*}

Running \textsc{Grad-MatchPB} for Librispeech is prohibitively expensive since the amount of memory required to store all the gradients would exceed the memory size of available commercial GPUs as described in Section \ref{need}. To address this, we compare Phone Error (PER) on the TIMIT Phone recognition dataset \cite{garofolo1993timit} (containing 3680 utterances with 630 speakers) for all the methods. 
 
Table \ref{tab:pb_vs_pgm} shows WER obtained with \dsys{}, \textsc{Grad-MatchPB} \random{} and other subset selection baselines such as \largeSmall{}, \largeOnly{}. For \dsys{} we use data partitioning $D=2$. We see that the WER of \dsys{} is slightly higher than that of \textsc{Grad-MatchPB}, as the error term  that \dsys{} minimises is a upper bound of error term minimised by \textsc{Grad-MatchPB} as discussed in section \ref{conn_ew}. However, \dsys{}'s WER is very close to that of \textsc{Grad-MatchPB}, indicating that the partitioning doesn’t deteriorate the bounds while allowing to scale for larger datasets and utilize multiple GPUs which allows \dsys{} to enjoy better speedups over \textsc{Grad-MatchPB}.

\textbf{Statistical Significance: } WER reductions using \dsys{} compared to the \random{} baseline are statistically significant at $p < 0.001$ using a matched pairs test.\footnote{https://github.com/talhanai/wer-sigtest}

\section{Conclusion}
We propose \dsys, a distributable data subset selection algorithm which avoids the need to load the entire dataset at a time, by constructing partial subsets from smaller data partitions. \dsys{} is an adaptive subset selection algorithm that improves the training time of the ASR models while maintaining low relative test error as compared to the ASR model trained with the entire dataset. This speed-up improves the efficiency 
of the training process and subsequently reduces the carbon footprint of training such models. 
Our approach  performs consistently better than \random{} baseline whilst providing good speed up, and robustness in the presence of noise. Although we test the method on the RNN-T model, we believe that similar results could be obtained for other ASR models and we leave that as future work.

\section*{Limitations}

In this paper we investigate the usefulness of subset selection algorithms for the ASR task for the first time on a popular RNN-Transducer ASR architecture which typically consumes vast volume ($\sim$40000 hours of labelled audio and more) of training data. At such industrial scale, the overhead of \dsys{} for gradient matching over the entire training set would limit the utility of the algorithm. These practical considerations warrant more careful design of the subset selection algorithms so as to scale well with such huge workloads. We also limit our results to the RNN-T architecture and believe that the results also hold for other less popular architecture by taking gradients of the last few layers. While we show the efficient training for the ASR task, we believe a similar study should be carried out for the self-supervised pre-training approaches.

\section*{Ethics Statement}
In this work we present a gradient matching based data subset selection algorithm for compute efficient and robust ASR model training. Since we do not modify any existing speech architecture or propose new benchmarks, but provide a mechanism for faster training of such models, we see no new ethical concerns arising from our work.

\section{Acknowledgements}
Durga Sivasubramanian is supported by the Prime Minister’s Research Fellowship. The authors gratefully acknowledge the support from IBM Research, specifically the IBM AI Horizon Networks-IIT Bombay initiative.  Ganesh Ramakrishnan is grateful to the IIT Bombay Institute Chair Professorship for
their support and sponsorship. Rishabh Iyer acknowledges support from NSF Grant Number IIS-2106937, a gift from Google Research, and an Adobe Data Science Research award.

\bibliography{mybib}

\begin{thebibliography}{37}
\expandafter\ifx\csname natexlab\endcsname\relax\def\natexlab#1{#1}\fi

\bibitem[{Chan et~al.(2016)Chan, Jaitly, Le, and Vinyals}]{chan2016listen}
William Chan, Navdeep Jaitly, Quoc Le, and Oriol Vinyals. 2016.
\newblock Listen, attend and spell: A neural network for large vocabulary
  conversational speech recognition.
\newblock In \emph{2016 IEEE international conference on acoustics, speech and
  signal processing (ICASSP)}, pages 4960--4964. IEEE.

\bibitem[{Clarkson(2010)}]{clarkson2010coresets}
Kenneth~L Clarkson. 2010.
\newblock Coresets, sparse greedy approximation, and the frank-wolfe algorithm.
\newblock \emph{ACM Transactions on Algorithms (TALG)}, 6(4):1--30.

\bibitem[{Coleman et~al.(2020)Coleman, Yeh, Mussmann, Mirzasoleiman, Bailis,
  Liang, Leskovec, and Zaharia}]{coleman2020selection}
Cody Coleman, Christopher Yeh, Stephen Mussmann, Baharan Mirzasoleiman, Peter
  Bailis, Percy Liang, Jure Leskovec, and Matei Zaharia. 2020.
\newblock \href {http://arxiv.org/abs/1906.11829} {Selection via proxy:
  Efficient data selection for deep learning}.

\bibitem[{Elenberg et~al.(2018)Elenberg, Khanna, Dimakis, Negahban
  et~al.}]{elenberg2018restricted}
Ethan~R Elenberg, Rajiv Khanna, Alexandros~G Dimakis, Sahand Negahban, et~al.
  2018.
\newblock Restricted strong convexity implies weak submodularity.
\newblock \emph{The Annals of Statistics}, 46(6B):3539--3568.

\bibitem[{Garofolo(1993)}]{garofolo1993timit}
John~S Garofolo. 1993.
\newblock Timit acoustic phonetic continuous speech corpus.
\newblock \emph{Linguistic Data Consortium, 1993}.

\bibitem[{Graves(2012)}]{graves2012sequence}
Alex Graves. 2012.
\newblock Sequence transduction with recurrent neural networks.
\newblock \emph{arXiv preprint arXiv:1211.3711}.

\bibitem[{Graves et~al.(2006)Graves, Fern{\'a}ndez, Gomez, and
  Schmidhuber}]{graves2006connectionist}
Alex Graves, Santiago Fern{\'a}ndez, Faustino Gomez, and J{\"u}rgen
  Schmidhuber. 2006.
\newblock Connectionist temporal classification: labelling unsegmented sequence
  data with recurrent neural networks.
\newblock In \emph{Proceedings of the 23rd international conference on Machine
  learning}, pages 369--376.

\bibitem[{Graves et~al.(2013)Graves, Mohamed, and Hinton}]{graves2013speech}
Alex Graves, Abdel-rahman Mohamed, and Geoffrey Hinton. 2013.
\newblock Speech recognition with deep recurrent neural networks.
\newblock In \emph{2013 IEEE international conference on acoustics, speech and
  signal processing}, pages 6645--6649. Ieee.

\bibitem[{Gulati et~al.(2020)Gulati, Qin, Chiu, Parmar, Zhang, Yu, Han, Wang,
  Zhang, Wu et~al.}]{gulati2020conformer}
Anmol Gulati, James Qin, Chung-Cheng Chiu, Niki Parmar, Yu~Zhang, Jiahui Yu,
  Wei Han, Shibo Wang, Zhengdong Zhang, Yonghui Wu, et~al. 2020.
\newblock Conformer: Convolution-augmented transformer for speech recognition.
\newblock \emph{arXiv preprint arXiv:2005.08100}.

\bibitem[{Hannun et~al.(2019)Hannun, Lee, Xu, and
  Collobert}]{hannun2019sequence}
Awni Hannun, Ann Lee, Qiantong Xu, and Ronan Collobert. 2019.
\newblock Sequence-to-sequence speech recognition with time-depth separable
  convolutions.
\newblock \emph{arXiv preprint arXiv:1904.02619}.

\bibitem[{Har-Peled and Mazumdar(2004)}]{har2004coresets}
Sariel Har-Peled and Soham Mazumdar. 2004.
\newblock On coresets for k-means and k-median clustering.
\newblock In \emph{Proceedings of the thirty-sixth annual ACM symposium on
  Theory of computing}, pages 291--300.

\bibitem[{He et~al.(2016)He, Zhang, Ren, and Sun}]{he2016deep}
Kaiming He, Xiangyu Zhang, Shaoqing Ren, and Jian Sun. 2016.
\newblock Deep residual learning for image recognition.
\newblock In \emph{Proceedings of the IEEE conference on computer vision and
  pattern recognition}, pages 770--778.

\bibitem[{He et~al.(2019)He, Sainath, Prabhavalkar, McGraw, Alvarez, Zhao,
  Rybach, Kannan, Wu, Pang et~al.}]{he2019streaming}
Yanzhang He, Tara~N Sainath, Rohit Prabhavalkar, Ian McGraw, Raziel Alvarez,
  Ding Zhao, David Rybach, Anjuli Kannan, Yonghui Wu, Ruoming Pang, et~al.
  2019.
\newblock Streaming end-to-end speech recognition for mobile devices.
\newblock In \emph{ICASSP 2019-2019 IEEE International Conference on Acoustics,
  Speech and Signal Processing (ICASSP)}, pages 6381--6385. IEEE.

\bibitem[{Hrinchuk et~al.(2020)Hrinchuk, Popova, and
  Ginsburg}]{hrinchuk2020correction}
Oleksii Hrinchuk, Mariya Popova, and Boris Ginsburg. 2020.
\newblock Correction of automatic speech recognition with transformer
  sequence-to-sequence model.
\newblock In \emph{ICASSP 2020-2020 IEEE International Conference on Acoustics,
  Speech and Signal Processing (ICASSP)}, pages 7074--7078. IEEE.

\bibitem[{Kannan et~al.(2018)Kannan, Wu, Nguyen, Sainath, Chen, and
  Prabhavalkar}]{kannan2018analysis}
Anjuli Kannan, Yonghui Wu, Patrick Nguyen, Tara~N Sainath, Zhijeng Chen, and
  Rohit Prabhavalkar. 2018.
\newblock An analysis of incorporating an external language model into a
  sequence-to-sequence model.
\newblock In \emph{2018 IEEE International Conference on Acoustics, Speech and
  Signal Processing (ICASSP)}, pages 1--5828. IEEE.

\bibitem[{Kaushal et~al.(2019)Kaushal, Iyer, Kothawade, Mahadev, Doctor, and
  Ramakrishnan}]{kaushal2019learning}
Vishal Kaushal, Rishabh Iyer, Suraj Kothawade, Rohan Mahadev, Khoshrav Doctor,
  and Ganesh Ramakrishnan. 2019.
\newblock Learning from less data: A unified data subset selection and active
  learning framework for computer vision.
\newblock In \emph{2019 IEEE Winter Conference on Applications of Computer
  Vision (WACV)}, pages 1289--1299. IEEE.

\bibitem[{Killamsetty et~al.(2021{\natexlab{a}})Killamsetty, S, Ramakrishnan,
  De, and Iyer}]{pmlr-v139-killamsetty21a}
Krishnateja Killamsetty, Durga S, Ganesh Ramakrishnan, Abir De, and Rishabh
  Iyer. 2021{\natexlab{a}}.
\newblock Grad-match: Gradient matching based data subset selection for
  efficient deep model training.
\newblock In \emph{Proceedings of the 38th International Conference on Machine
  Learning}, volume 139 of \emph{Proceedings of Machine Learning Research},
  pages 5464--5474. PMLR.

\bibitem[{Killamsetty et~al.(2021{\natexlab{b}})Killamsetty, Sivasubramanian,
  Ramakrishnan, and Iyer}]{killamsetty2021glister}
Krishnateja Killamsetty, Durga Sivasubramanian, Ganesh Ramakrishnan, and
  Rishabh Iyer. 2021{\natexlab{b}}.
\newblock Glister: Generalization based data subset selection for efficient and
  robust learning.
\newblock \emph{In AAAI}.

\bibitem[{Krizhevsky(2009)}]{Krizhevsky09learningmultiple}
Alex Krizhevsky. 2009.
\newblock Learning multiple layers of features from tiny images.
\newblock Technical report.

\bibitem[{Liu et~al.(2017)Liu, Iyer, Kirchhoff, and
  Bilmes}]{liu2017svitchboard}
Yuzong Liu, Rishabh Iyer, Katrin Kirchhoff, and Jeff Bilmes. 2017.
\newblock Svitchboard-ii and fisver-i: Crafting high quality and low complexity
  conversational english speech corpora using submodular function optimization.
\newblock \emph{Computer Speech \& Language}, 42:122--142.

\bibitem[{Mirzasoleiman et~al.(2020)Mirzasoleiman, Bilmes, and
  Leskovec}]{mirzasoleiman2020coresets}
Baharan Mirzasoleiman, Jeff Bilmes, and Jure Leskovec. 2020.
\newblock \href {http://arxiv.org/abs/1906.01827} {Coresets for data-efficient
  training of machine learning models}.

\bibitem[{Natarajan(1995)}]{natarajan1995sparse}
Balas~Kausik Natarajan. 1995.
\newblock Sparse approximate solutions to linear systems.
\newblock \emph{SIAM journal on computing}, 24(2):227--234.

\bibitem[{Panayotov et~al.(2015)Panayotov, Chen, Povey, and
  Khudanpur}]{panayotov2015librispeech}
Vassil Panayotov, Guoguo Chen, Daniel Povey, and Sanjeev Khudanpur. 2015.
\newblock Librispeech: an asr corpus based on public domain audio books.
\newblock In \emph{2015 IEEE international conference on acoustics, speech and
  signal processing (ICASSP)}, pages 5206--5210. IEEE.

\bibitem[{Parcollet and Ravanelli(2021)}]{parcollet2021energy}
Titouan Parcollet and Mirco Ravanelli. 2021.
\newblock The energy and carbon footprint of training end-to-end speech
  recognizers.
\newblock \emph{Interspeech}.

\bibitem[{Rao et~al.(2017)Rao, Sak, and Prabhavalkar}]{rao2017exploring}
Kanishka Rao, Ha{\c{s}}im Sak, and Rohit Prabhavalkar. 2017.
\newblock Exploring architectures, data and units for streaming end-to-end
  speech recognition with rnn-transducer.
\newblock In \emph{2017 IEEE Automatic Speech Recognition and Understanding
  Workshop (ASRU)}, pages 193--199. IEEE.

\bibitem[{Ravanelli et~al.(2021)Ravanelli, Parcollet, Plantinga, Rouhe,
  Cornell, Lugosch, Subakan, Dawalatabad, Heba, Zhong
  et~al.}]{ravanelli2021speechbrain}
Mirco Ravanelli, Titouan Parcollet, Peter Plantinga, Aku Rouhe, Samuele
  Cornell, Loren Lugosch, Cem Subakan, Nauman Dawalatabad, Abdelwahab Heba,
  Jianyuan Zhong, et~al. 2021.
\newblock Speechbrain: A general-purpose speech toolkit.
\newblock \emph{arXiv preprint arXiv:2106.04624}.

\bibitem[{Sainath et~al.(2020)Sainath, He, Li, Narayanan, Pang, Bruguier,
  Chang, Li, Alvarez, Chen et~al.}]{sainath2020streaming}
Tara~N Sainath, Yanzhang He, Bo~Li, Arun Narayanan, Ruoming Pang, Antoine
  Bruguier, Shuo-yiin Chang, Wei Li, Raziel Alvarez, Zhifeng Chen, et~al. 2020.
\newblock A streaming on-device end-to-end model surpassing server-side
  conventional model quality and latency.
\newblock In \emph{ICASSP 2020-2020 IEEE International Conference on Acoustics,
  Speech and Signal Processing (ICASSP)}, pages 6059--6063. IEEE.

\bibitem[{Saon et~al.(2020)Saon, T{\"u}ske, and Audhkhasi}]{saon2020alignment}
George Saon, Zolt{\'a}n T{\"u}ske, and Kartik Audhkhasi. 2020.
\newblock Alignment-length synchronous decoding for rnn transducer.
\newblock In \emph{ICASSP 2020-2020 IEEE International Conference on Acoustics,
  Speech and Signal Processing (ICASSP)}, pages 7804--7808. IEEE.

\bibitem[{Saon et~al.(2021)Saon, T{\"u}ske, Bolanos, and
  Kingsbury}]{saon2021advancing}
George Saon, Zolt{\'a}n T{\"u}ske, Daniel Bolanos, and Brian Kingsbury. 2021.
\newblock Advancing rnn transducer technology for speech recognition.
\newblock In \emph{ICASSP 2021-2021 IEEE International Conference on Acoustics,
  Speech and Signal Processing (ICASSP)}, pages 5654--5658. IEEE.

\bibitem[{Schwartz et~al.(2020)Schwartz, Dodge, Smith, and
  Etzioni}]{schwartz2020green}
Roy Schwartz, Jesse Dodge, Noah~A Smith, and Oren Etzioni. 2020.
\newblock Green ai.
\newblock \emph{Communications of the ACM}, 63(12):54--63.

\bibitem[{Sharir et~al.(2020)Sharir, Peleg, and Shoham}]{sharir2020cost}
Or~Sharir, Barak Peleg, and Yoav Shoham. 2020.
\newblock The cost of training nlp models: A concise overview.
\newblock \emph{arXiv preprint arXiv:2004.08900}.

\bibitem[{Strubell et~al.(2019)Strubell, Ganesh, and
  McCallum}]{strubell2019energy}
Emma Strubell, Ananya Ganesh, and Andrew McCallum. 2019.
\newblock Energy and policy considerations for deep learning in nlp.
\newblock \emph{arXiv preprint arXiv:1906.02243}.

\bibitem[{Vaswani et~al.(2017)Vaswani, Shazeer, Parmar, Uszkoreit, Jones,
  Gomez, Kaiser, and Polosukhin}]{vaswani2017attention}
Ashish Vaswani, Noam Shazeer, Niki Parmar, Jakob Uszkoreit, Llion Jones,
  Aidan~N Gomez, {\L}ukasz Kaiser, and Illia Polosukhin. 2017.
\newblock Attention is all you need.
\newblock \emph{Advances in neural information processing systems}, 30.

\bibitem[{Watanabe et~al.(2017)Watanabe, Hori, Kim, Hershey, and
  Hayashi}]{watanabe2017hybrid}
Shinji Watanabe, Takaaki Hori, Suyoun Kim, John~R Hershey, and Tomoki Hayashi.
  2017.
\newblock Hybrid ctc/attention architecture for end-to-end speech recognition.
\newblock \emph{IEEE Journal of Selected Topics in Signal Processing},
  11(8):1240--1253.

\bibitem[{Wei et~al.(2014)Wei, Iyer, and Bilmes}]{wei2014fast}
Kai Wei, Rishabh Iyer, and Jeff Bilmes. 2014.
\newblock Fast multi-stage submodular maximization.
\newblock In \emph{International conference on machine learning}, pages
  1494--1502. PMLR.

\bibitem[{Wolf et~al.(2019)Wolf, Debut, Sanh, Chaumond, Delangue, Moi, Cistac,
  Rault, Louf, Funtowicz et~al.}]{wolf2019huggingface}
Thomas Wolf, Lysandre Debut, Victor Sanh, Julien Chaumond, Clement Delangue,
  Anthony Moi, Pierric Cistac, Tim Rault, R{\'e}mi Louf, Morgan Funtowicz,
  et~al. 2019.
\newblock Huggingface's transformers: State-of-the-art natural language
  processing.
\newblock \emph{arXiv preprint arXiv:1910.03771}.

\bibitem[{Zhao et~al.(2021)Zhao, Xue, Li, Wei, He, and
  Gong}]{zhao2021addressing}
Rui Zhao, Jian Xue, Jinyu Li, Wenning Wei, Lei He, and Yifan Gong. 2021.
\newblock On addressing practical challenges for rnn-transducer.
\newblock \emph{arXiv preprint arXiv:2105.00858}.

\end{thebibliography}
\bibliographystyle{acl_natbib}

\newpage
\appendix
\section{Connections between \dsys{} and \textsc{Grad-MatchPB}}\label{sec:proof}

\begin{lemma}  (triangle inequality). Let ${v_1, . . . , v_{\tau} }$ be $\tau$ vectors in $\mathbb{R}^d$. Then the following is true:
\begin{align}
    \|\sum_{i=1}^{\tau}v_i\| \geq  \sum_{i=1}^{\tau}\|v_i\|
\label{relax_tri}    
\end{align}
\end{lemma}

\begin{corollary}

    Following inequality holds between the objectives of \dsys{} and \textsc{Grad-MatchPB} 
    \begin{equation*}
    \begin{aligned}
    \mathbb{E}(\mbox{E}_\lambda(\mathbf{w}^t_{d^p}, \Xcal^t_{d^p}, L^{d^p}_T,& \nabla_{\theta} L^{d^p}_T, \theta^t)) \\& \geq \mbox{E}_\lambda(\mathbf{w}^t, \Xcal^t, L_T, L^{b_n}_T, \theta^t))
    \end{aligned}
\end{equation*}

and 

\begin{equation*}
    \begin{aligned}
    \mathbb{E}(\mbox{E}_\lambda(\mathbf{w}^t_{d^p}, \Xcal^t_{d^p}, L_V,& \nabla_{\theta} L^{d^p}_T, \theta^t)) \\& \geq \mbox{E}_\lambda(\mathbf{w}^t, \Xcal^t, L_V, L^{b_n}_T, \theta^t))
    \end{aligned}
\end{equation*}
    \label{cor:mean}
\end{corollary}

\textbf{Proof.}

Using the triangle inequality, 

\begin{equation*}
    \begin{aligned}
      \sum_{i=p}^D & (\lVert \sum_{i \in \Xcal^t_{d^p}} \mathbf{w}^t_{id^p} \nabla_{\theta} L^{d^pB_i}_T - \nabla_{\theta} L^{d^p}_T(\theta^t) \rVert \\& + \lambda \lVert \mathbf{w}^t_{d^p} \rVert^2 ) \\ &\geq \lVert \sum_{i=p}^D (\sum_{i \in \Xcal^t_{d^p}} \mathbf{w}^t_{id^p} \nabla_{\theta} L^{d^pB_i}_T - \nabla_{\theta} L^{d^p}_T(\theta^t)) \rVert \\ &+ \lambda \lVert \sum_{i=p}^D \mathbf{w}^t_{d^p} \rVert^2 
    \end{aligned}
\end{equation*}

We divide both sides by $D$,

\begin{equation*}
    \begin{aligned}
      \frac{1}{D}\sum_{i=p}^D & (\lVert \sum_{i \in \Xcal^t_{d^p}} \mathbf{w}^t_{id^p} \nabla_{\theta} L^{d^pB_i}_T - \nabla_{\theta} L^{d^p}_T(\theta^t) \rVert \\ &+ \lambda \lVert \mathbf{w}^t_{d^p} \rVert^2 ) \\ &\geq \lVert \sum_{i=p}^D (\sum_{i \in \Xcal^t_{d^p}} \frac{\mathbf{w}^t_{id^p}}{D} \nabla_{\theta} L^{d^pB_i}_T) \\& - \frac{\sum_{i=p}^D (\nabla_{\theta} L^{d^p}_T(\theta^t))}{D} \rVert + \lambda \lVert \frac{\sum_{i=p}^D \mathbf{w}^t_{d^p}}{D} \rVert^2 
    \end{aligned}
\end{equation*}

\begin{equation*}
    \begin{aligned}
      \mathbb{E}(\mbox{E}_\lambda(\mathbf{w}^t_{d^p} &, \Xcal^t_{d^p}, L^{d^p}_T, \nabla_{\theta} L^{d^p}_T, \theta^t)) \\ &\geq \lVert \sum_{i=p}^D (\sum_{i \in \Xcal^t_{d^p}} \frac{\mathbf{w}^t_{id^p}}{D} \nabla_{\theta} L^{d^pB_i}_T) \\& - \frac{\sum_{i=p}^D (\nabla_{\theta} L^{d^p}_T(\theta^t))}{D} \rVert + \lambda \lVert \frac{\sum_{i=p}^D \mathbf{w}^t_{d^p}}{D} \rVert^2 
    \end{aligned}
\end{equation*}

Since $L_T = \mathbb{E}(L^{d^p}_T)$ and therefore $\mathbb{E}(\sum_{i=p}^D (\sum_{i \in \Xcal^t_{d^p}} \frac{\mathbf{w}^t_{id^p}}{D} \nabla_{\theta} L^{d^pB_i}_T)) = \sum_{i \in \Xcal^t}\mathbf{w}^t_{i} \nabla_{\theta} L_T^{B_i}(\theta^t)$ as they are obtained via gradient matching, 

\begin{equation*}
    \begin{aligned}
      \mathbb{E}(\mbox{E}_\lambda(\mathbf{w}^t_{d^p} &, \Xcal^t_{d^p}, L^{d^p}_T, \nabla_{\theta} L^{d^p}_T, \theta^t)) \\ &\geq \lVert \sum_{i \in \Xcal^t}\mathbf{w}^t_{i} \nabla_{\theta} L_T^{B_i}(\theta^t) - \nabla_{\theta} L_T(\theta^t) \rVert \\& + \lambda \lVert \mathbf{w}^t \rVert^2
    \end{aligned}
\end{equation*}

\begin{equation*}
    \begin{aligned}
    \mathbb{E}(\mbox{E}_\lambda(\mathbf{w}^t_{d^p}, \Xcal^t_{d^p}, L^{d^p}_T,& \nabla_{\theta} L^{d^p}_T, \theta^t)) \\& \geq \mbox{E}_\lambda(\mathbf{w}^t, \Xcal^t, L_T, L^{b_n}_T, \theta^t))
    \end{aligned}
\end{equation*}

\end{document}